\documentclass[letterpaper]{article} 
\usepackage{aaai2026}  
\usepackage{times}  
\usepackage{helvet}  
\usepackage{courier}  
\usepackage[hyphens]{url}  
\usepackage{graphicx} 
\usepackage{natbib}  
\usepackage{caption} 
\usepackage{algorithm}
\usepackage{algorithmic}

\usepackage{newfloat}
\usepackage{listings}
\DeclareCaptionStyle{ruled}{labelfont=normalfont,labelsep=colon,strut=off} 
\floatstyle{ruled}
\newfloat{listing}{tb}{lst}{}
\floatname{listing}{Listing}
\title{COACH: Collaborative Agents for Contextual Highlighting \\ A Multi-Agent Framework for Sports Video Analysis}
\author{
    Tsz-To Wong\textsuperscript{\rm 1},  
    Ching-Chun Huang\textsuperscript{\rm 1}\thanks{Corresponding author.}
    Hong-Han Shuai\textsuperscript{\rm 1}\footnotemark[1]
}
\affiliations{
    \textsuperscript{\rm 1}National Yang Ming Chiao Tung University\\
    No. 1001, Daxue Rd., East Dist., Hsinchu City 300, Taiwan\\
    tsztowong.cs13@nycu.edu.tw, chingchun@cs.nycu.edu.tw, hhshuai@nycu.edu.tw
}

\usepackage{bibentry}

\begin{document}

\maketitle

\begin{abstract}
Intelligent sports video analysis demands a comprehensive understanding of temporal context, from micro-level actions to macro-level game strategies. Existing end-to-end models often struggle with this temporal hierarchy, offering solutions that lack generalization, incur high development costs for new tasks, and suffer from poor interpretability. To overcome these limitations, we propose a reconfigurable Multi-Agent System (MAS) as a foundational framework for sports video understanding. In our system, each agent functions as a distinct "cognitive tool" specializing in a specific aspect of analysis. The system's architecture is not confined to a single temporal dimension or task. By leveraging iterative invocation and flexible composition of these agents, our framework can construct adaptive pipelines for both short-term analytic reasoning (e.g., Rally QA) and long-term generative summarization (e.g., match summaries). We demonstrate the adaptability of this framework using two representative tasks in badminton analysis, showcasing its ability to bridge fine-grained event detection and global semantic organization. This work presents a paradigm shift towards a flexible, scalable, and interpretable system for robust, cross-task sports video intelligence.
\end{abstract}

\begin{links}
    \link{Project Homepage}{https://aiden1020.github.io/COACH-project-page/}
\end{links}
\section{Introduction}

Sports video analysis goes beyond recognizing actions or detecting events. It aims to understand what happened, why it happened, and how it developed over time.~\cite{xiao2021next}

Games like badminton, with their fast-paced rallies and continuous player interactions, naturally form a multi-level timeline: micro-level (strokes and rallies), mid-level (sets), and macro-level (complete matches and overall tactics).~\cite{li2025sportsqalargescalevideoquestion} Such temporal complexity requires a system capable of multi-scale reasoning. A good analysis system therefore needs both detailed accuracy for local actions and global understanding of long-term strategies and flow.

Most existing systems rely on end-to-end single-model training~\cite{li2023blip, lin2023video,li2023videochat}, where each model is built for one specific task, such as Video Question Answering or Video Summarization. This design leads to three key problems:

\begin{enumerate}

\item \textbf{High Redundant Cost:} Each application requires a separate model and training pipeline, since components cannot be reused across different objectives, reducing scalability.
\item \textbf{Locked to a Single Temporal Scale:} End-to-end models are trained for one granularity (e.g., rally-level) and cannot transfer knowledge to other scales (e.g., game-level), limiting cross-level understanding.

\item \textbf{Opaque Reasoning:} The decision process is hidden, making it difficult to interpret, verify, or integrate new reasoning modules.

\end{enumerate}

To address these limitations, we propose a Multi-Agent System as a flexible foundation for sports video understanding. Each agent functions as an independent cognitive tool, responsible for a specific type of reasoning, such as player tracking, stroke recognition, rally segmentation, or tactical inference. This modular design allows the system to adapt and expand without rebuilding everything from scratch.

Our system is not tied to any single time scale or task type. It can perform both short-term reasoning (like answering a question about one rally) and long-term integration (like summarizing an entire match). This adaptability is achieved through iterative coordination and composable workflows, repeatedly invoking agents and combining their outputs into customized pipelines. In this way, the framework can flexibly switch between analytic reasoning and generative summarization as needed.

The research demonstrates a composable framework built around multi-agent collaboration. We validate the framework through two complementary applications: (1) Video QA, which focuses on fine-grained analytical understanding for single rally, and (2) Video Summarization, which focuses on global integration and narrative generation. Together, these tasks demonstrate the framework’s adaptability across different temporal levels and task objectives.
\begin{figure}
    \centering
    \includegraphics[width=1\linewidth]{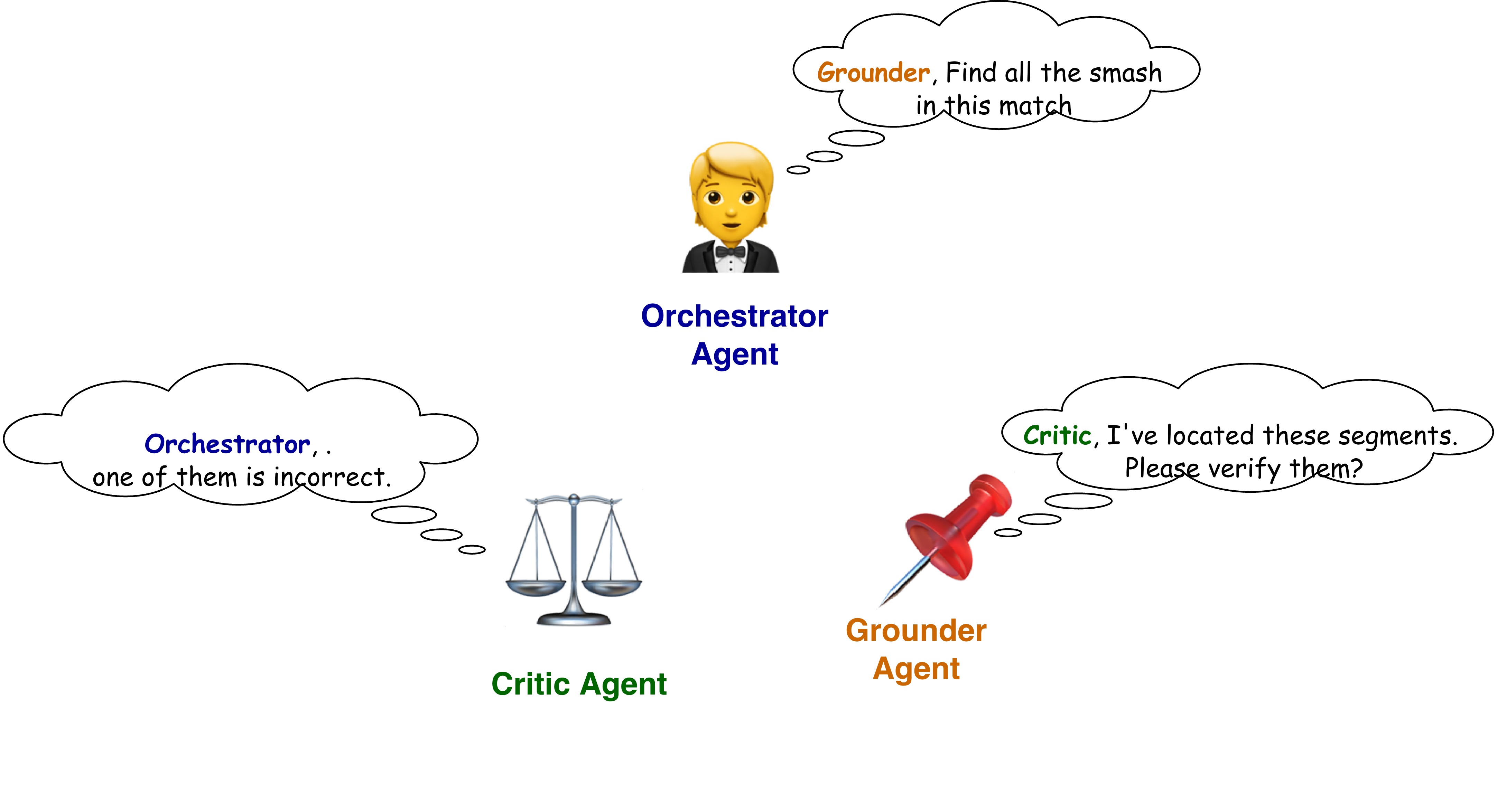}
    \caption{A high-level conceptual diagram illustrating the collaborative interaction between agents.}
    \label{fig:placeholder}
\end{figure}
\begin{figure}
    \centering
    \includegraphics[width=1\linewidth]{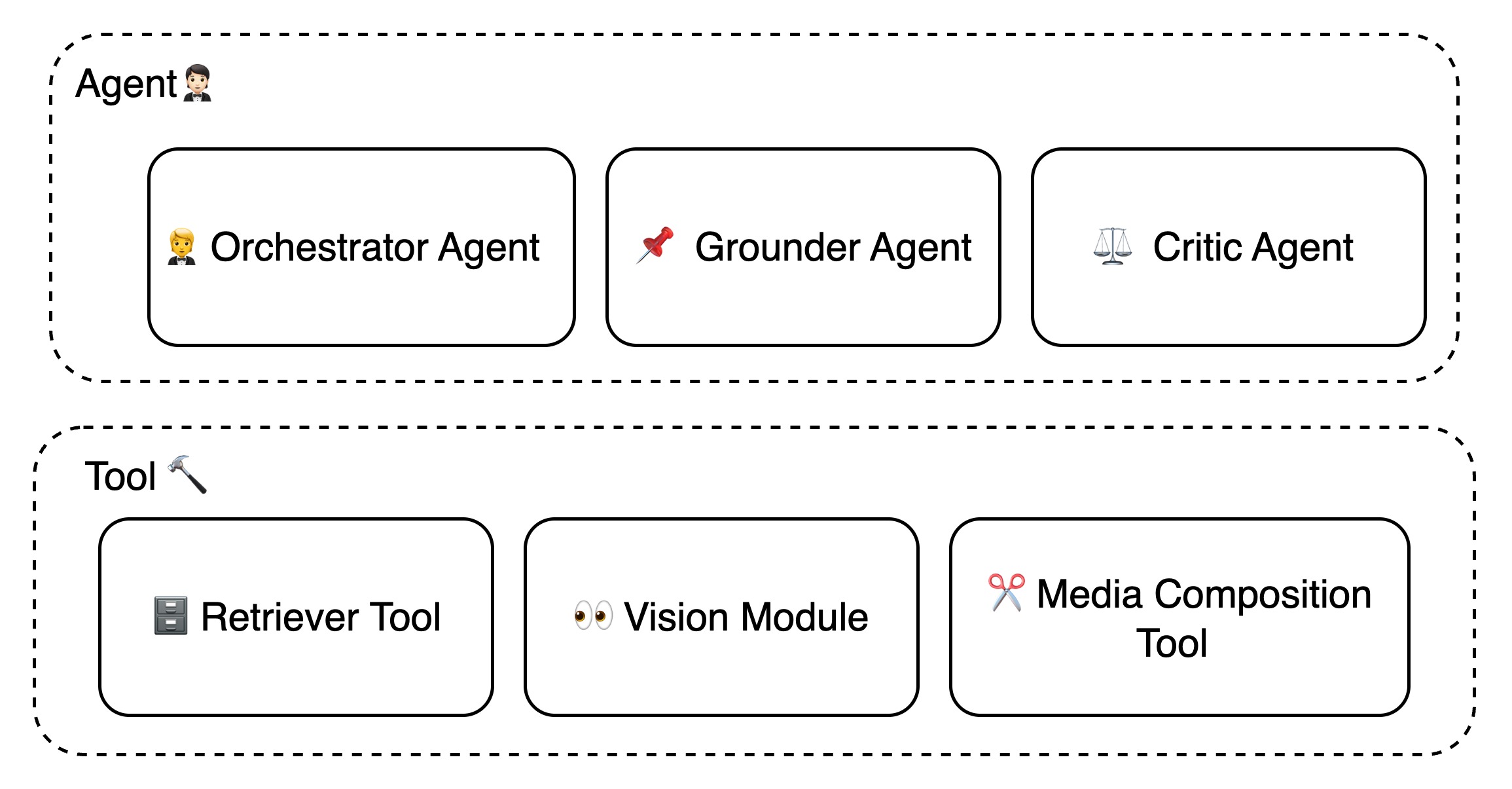}
    \caption{Shared components library}
    \label{fig:Shared components library}
\end{figure}
\section{Related Work}

\subsection{Video Understanding}

Recent advances in Video-Language Models (VLMs), such as BLIP-2~\cite{li2023blip}, Video-LLaVA~\cite{lin2023video}, and VideoChat~\cite{li2023videochat}, have shown remarkable progress in bridging visual and textual semantics for video understanding tasks. 
Although some recent models incorporate temporal fusion, most still process videos through fixed-length clips and lack explicit hierarchical temporal modeling. 
As a result, they struggle to capture the multi-level structure of long-form content such as sports matches, which unfold from strokes to rallies to sets~\cite{li2024sports, liu2025f}. 
Consequently, their reasoning remains confined to local event recognition rather than global narrative understanding. 
This imbalance is highlighted by sports video understanding datasets~\cite{shao2020finegym, xiao2021next}, 
showing that models capable of describing \textit{what} happens 
often fail to explain \textit{why} it happens due to missing causal and temporal dependencies.
These limitations motivate a more structured, system-level approach to temporal reasoning.

\subsection{Modular and Multi-Agent Frameworks}

Beyond temporal scalability, the monolithic design of end-to-end models restricts their reusability and coordination across applications. 
To overcome this issue, modular and multi-agent frameworks have emerged as promising alternatives. 
In the language domain, systems such as HuggingGPT~\cite{shen2023hugginggpt} employ a large language model (LLM) as a central controller to delegate tasks to specialized sub-models, while MetaGPT~\cite{hong2023metagpt} introduces role-based agents coordinated through Standard Operating Procedures (SOPs). 
In the vision domain, AgentOrchestra~\cite{zhang2025agentorchestra} demonstrates hierarchical orchestration and dynamic workflow composition, improving generalization and plug-and-play capabilities. 
However, these frameworks mainly focus on static task coordination and lack temporal reasoning across multiple levels of abstraction.

\subsection{Instruction Tuning and Role Specialization}
Instruction tuning offers another path toward controllable and interpretable agent behavior. 
Early multimodal works~\cite{dai2023instructblip, liu2023visualinstructiontuning} established instructions as a universal interface for cross-modal reasoning. 
Later studies introduced role-based prompting and self-refinement~\cite{madaan2023self, zhou2023language}, allowing a single model to simulate collaborative reasoning among roles. 
More recent approaches, such as AgentTuning~\cite{zeng2023agenttuning} and AgentBank~\cite{song2024agentbank}, extend this concept by fine-tuning agents on role-specific instructional data, embedding skills such as planning or tool use. 
While these frameworks enable role specialization, they still train each agent independently, without mechanisms for joint optimization or shared feedback. 
Our work builds upon this direction by proposing a framework in which agents, such as the Orchestrator, Grounder, and Critic, are not only fine-tuned for specific roles but also learn to coordinate explicitly to form a verifiable and composable system for sports video understanding.
\section{Method}
\label{sec:method}
This section details our \textbf{COACH}: \textbf{CO}llaborative \textbf{A}gents for \textbf{C}ontextual \textbf{H}ighlighting framework. The methodology is presented as follows: We first introduce our core architectural foundation—a single shared backbone—and its system-level motivation. We then describe how this backbone operates at a macro-level (Orchestration) and a micro-level (Tuning), and finally define the components involved.

\subsection{Architectural Foundation: A Shared Backbone for Scalable Deployment}
Our methodology is built upon a foundational design choice: a \textbf{single shared backbone model} for all agents. This decision is motivated by system-level scalability and efficiency. A unified backbone, deployed via multi-GPU parallelism, can concurrently serve high-throughput batch inference (e.g., summarization) and high-concurrency requests (e.g., QA). This design also inherently provides global knowledge alignment by ensuring all agents operate within a unified representation space. The following sections detail the orchestration and specialization mechanisms that enable this single model to perform its diverse, multi-role functions.

\subsection{Intent-Driven Orchestration for Policy-Guided Collaboration}
Given this single-backbone architecture, the system coordinates tasks at a macro-level using policy-guided orchestration, which offers greater stability than the "online planning" paradigms employed in many traditional systems.

\subsubsection{Design: Intent Routing as Policy Selection}
In our framework, a "policy" is not a dynamic, step-by-step decision model. Instead, it is a pre-defined, goal-oriented collaboration plan, such as a Standard Operating Procedure (SOP), that specifies a fixed, reproducible sequence of agent and tool invocations.

The Orchestrator's core reasoning capability, trained via supervised imitation, lies in performing intent-to-strategy mapping. Given a user query (e.g., "What happened in rally 5?" vs. "Create a highlight reel"), its first action is to classify the semantic intent and select the corresponding pre-defined plan. This design is stable and reproducible because the collaboration flows are fixed, yet remains adaptive because the agents *within* that plan operate dynamically based on the context.

\subsubsection{Pre-defined Collaboration Strategies}
\begin{figure*}[t]
\centering
\includegraphics[width=0.9\textwidth]{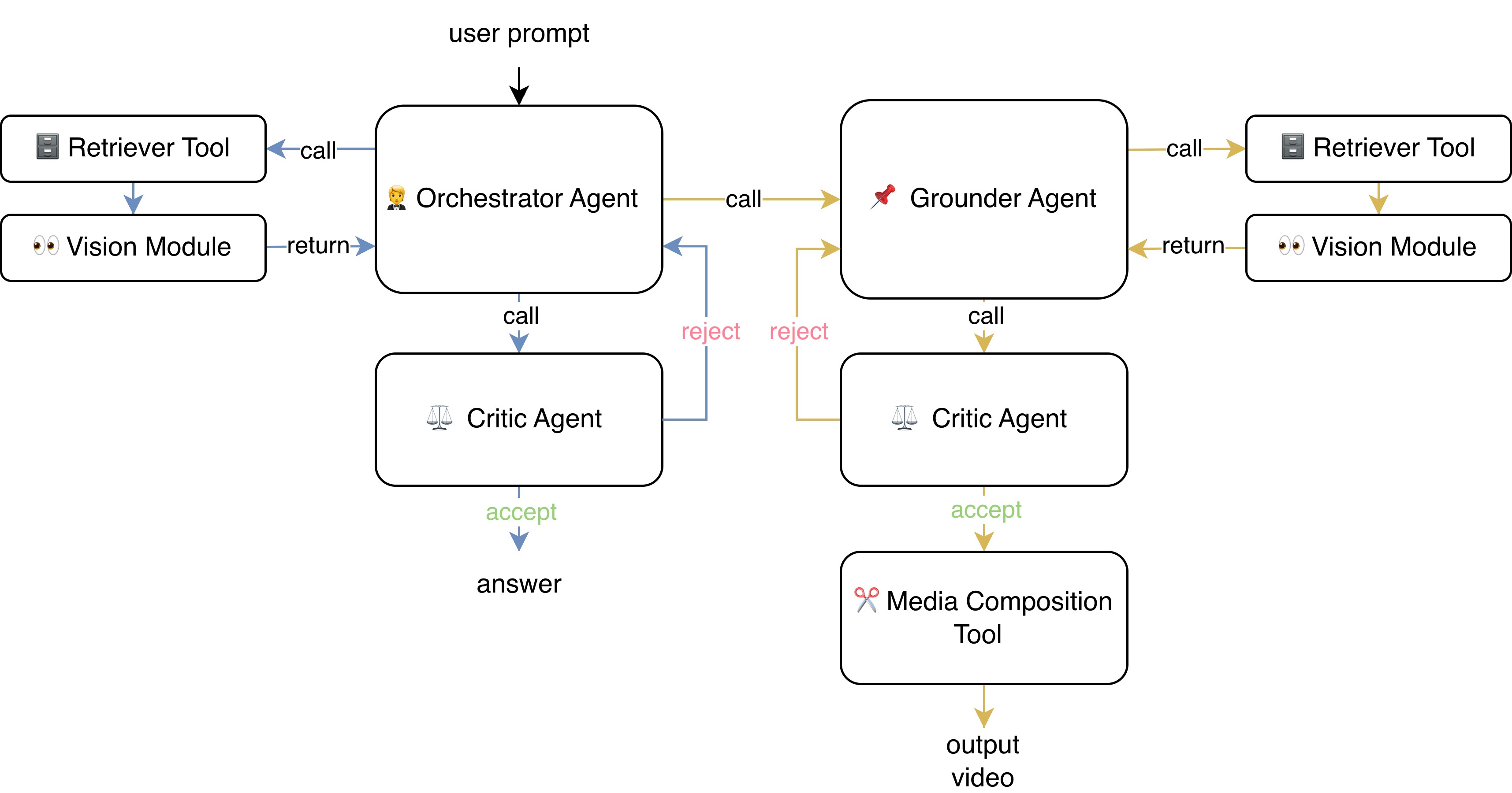} 
\caption{
The multi-agent collaboration workflow in COACH. Based on the user's prompt, the \texttt{Orchestrator Agent} acts as an intent router, initiating one of two distinct collaboration policies : 
\textbf{(Left)} The \textbf{Analytical Rally QA} pipeline, which uses an \texttt{Orchestrator-Critic} loop to verify evidence and generate a factual text answer.
\textbf{(Right)} The \textbf{Generative Video Summarization} pipeline, which uses a specialist \texttt{Grounder-Critic} loop to perform high-precision temporal localization, followed by the \texttt{Media Composition Tool} to assemble the final video output.
}
\label{fig2}
\end{figure*}

We implement two primary, pre-defined strategies that correspond to the main tasks of our system (as illustrated in Fig ~\ref{fig2}).

\textbf{Analytical Rally QA Strategy} is designed for precise, evidence-based factual reasoning. This flow typically involves the Orchestrator first calling the Retriever to find relevant evidence, then calling the Critic to verify the factual consistency of the information, and finally synthesizing the verified information into a conclusive answer.

\textbf{Generative Video Summarization Strategy} is designed for narrative, open-ended generation. This flow involves a more complex sequence: the Orchestrator first plans the summary structure, then calls the Grounder (often in a batch) to localize all required events. These events are then verified by the Critic, synthesized into a narrative script by the Orchestrator, and finally compiled into a video by the Media Composition tool.

\subsection{Multi-Role Specialization via Structured CoT Tuning}
This single-backbone, multi-strategy approach introduces the significant challenge of role conflict. We must enable functionally distinct capabilities, such as the 'complex planning' required by the Orchestrator and the 'suppressed reasoning' demanded by the Grounder—to coexist within the same model weights without task-level interference.

Our solution is \textbf{Structured CoT Tuning}. We mix multiple sets of role-specific, structured Chain-of-Thought (CoT) instruction datasets within the single shared backbone. This induces "multi-persona reasoning modes." Role-switching is achieved not by different weights, but by adhering to different, supervised CoT structural templates during inference.

We ensure clean behavioral separation through two primary mechanisms:

\subsubsection{Instruction Conditioning}
We use role-specific instruction prefixes, such as "You are a Grounder Agent responsible for temporal localization." This guides the model into a role-specific reasoning subspace during inference, thereby mitigating role interference.

\subsubsection{Role-Specific CoT Design}
We design distinct CoT templates (thinking patterns) for each agent, each optimized for its specific function. This is where the model's specialized behaviors are learned:
\begin{itemize} 
    \item \textbf{Orchestrator Agent:}
    It is trained to be the "strategist" with the highest degree of reasoning freedom. Its CoT pattern is a \textbf{Conditional Routing} mechanism. The template trains the model to first analyze the query's intent: 
    (1) For text-based tasks, it executes a single-step inference to answer directly from its internal knowledge. 
    (2) For video-based tasks, the CoT pattern guides the model to perform the high-level, multi-step visual reasoning (e.g., analyzing tactics, counting actions) required to synthesize the final answer.
    (3) For video summarization task, it starts to breakdown the high level query into multiple in domain queries and pass these queries to grounder agent for fine grained temporal localization. 
    \item \textbf{Grounder Agent:}
    The agent is trained as a high-precision "executor" role. Its CoT pattern is a rigid, non-reasoning, \textbf{Observe $\to$ Report structure}. The model is trained to strictly parse the instruction (e.g., "find all smashes") and report only the factual temporal locations (e.g., `[stroke 3, stroke 7]`). Crucially, the agent needs to report an empty set if no visual evidence is found, ensuring high stability and verifiability.

    \item \textbf{Critic Agent:}
    The agent's thinking pattern is designed to be \textbf{evaluative and oppositional}, operating in direct contrast to the generative nature of the \texttt{Orchestrator}. Its goal is to be an interpretable, "fact-checking engine." To achieve this, its CoT template follows a traceable, forensic structure: \textbf{"Analyze Assertion $\to$ Compare Evidence $\to$ Adjudicate Verdict"}. Unlike the \texttt{Orchestrator}, which reasons "forward" (What should I do?), the \texttt{Critic} reasons "backward" (Is this claim true?). It is trained to take a claim as input (e.g., "Stroke 5 was a smash") and rigorously compare it against the visual evidence before rendering a verifiable judgment. This oppositional structure is critical for mitigating factual errors and ensuring the system's overall reliability.
\end{itemize}


\subsection{Component and Tool Definitions}
For clarity, we formally define the agents and tools in Fig \ref{fig:Shared components library} that comprise our system:
\begin{itemize}
    \item \textbf{Core Agents:} Instantiated from the shared backbone.
    \begin{itemize}
        \item \textbf{Orchestrator Agent:} The central controller that performs intent routing and task-adaptive control.
        \item \textbf{Grounder Agent:} The high-precision execution agent for temporal localization.
        \item \textbf{Critic Agent:} The verification agent for fact-checking and consistency.
    \end{itemize}
    \item \textbf{Foundation Modules (Tools):}
    \begin{itemize}
        \item \textbf{Vision Module:} Converts video into semantic features.
        \item \textbf{Retriever Tool:} Provides video or knowledge retrieval.
        \item \textbf{Media Composition Tool:} Assembles video clips for summarization.
    \end{itemize}
\end{itemize}
\begin{table*}[t]
\centering
\caption{Performance on the \textbf{Analytical Rally QA} task. This demonstrates the superiority of the \textbf{COACH} pipeline over a generalist model for complex, high-level reasoning.}
\label{tab:rally_qa_results}
\resizebox{\textwidth}{!}{
\begin{tabular}{l|c|c|c|ccc|c}
\hline
\textbf{Model} & \textbf{Action Class.} & \textbf{Action Count} & \textbf{Summarisation} & \multicolumn{3}{c|}{\textbf{Temporal Localization}} & \textbf{Knowledge QA}\\
 & (EM \%) $\uparrow$ & (EM \%) $\uparrow$ & (ROUGE-L) $\uparrow$ & (Hit@1 \%) $\uparrow$ & (EM \%) $\uparrow$ & (F1 \%) $\uparrow$ & (ROUGE-L) $\uparrow$ \\
\hline
\multicolumn{8}{l}{\textit{Main Comparison: Specialist vs. Generalist}} \\
\hline
Gemini 2.5 Pro & 24.20 & 37.60 & 23.55 & 18.12 & 7.04 & 13.73 & \textbf{29.76} \\
\textbf{COACH (w/ Critic)} & \textbf{85.60} & \textbf{79.20} & \textbf{33.56} & \textbf{76.66} & \textbf{63.97} & \textbf{76.21} & 27.40 \\
\hline
\multicolumn{8}{l}{\textit{Ablation Studies: Impact of Architecture Choices}} \\
\hline
COACH (w/o Critic) & 82.20 & 79.60 & 32.24 & 76.94 & 61.09 & 73.63 & 27.40 \\
\hline
\end{tabular}

}
\end{table*}
\section{Experiments}

\subsection{Dataset Construction}

We constructed a dataset named \textbf{COACH-Data}, to serve as the primary source for both training our agents and evaluating their performance. This dataset is built from 22 badminton matches, comprising approximately 19,000 annotated strokes. We observed that existing video understanding datasets either lack this required fine-grained temporal data or do not possess the complex causal reasoning queries needed for tactical analysis. Our dataset is thus composed of two distinct sub-tasks, defined by their source and purpose.

\subsubsection{Dataset Analysis}

\paragraph{Video-Based QA }
To train and test fine-grained, video-based reasoning, we built a sub-dataset based on the \textbf{ShuttleSet} ~\cite{wang2023shuttlesethumanannotatedstrokelevelsingles} annotations. We selected ShuttleSet for its high-quality, human-annotated stroke-level labels. However, ShuttleSet itself is a dataset for stroke classification and low level analysis. We introduce a new task format by building a dataset of complex, multi-step questions. This dataset is designed to cover a range of hierarchical reasoning skills, including \textbf{action classification} (e.g., "What shot is stroke 5?"), \textbf{action counting} (e.g., "How many smashes occurred?"), \textbf{temporal localization} (e.g., "When does the player serve?"), and complex \textbf{summarization \& analysis} (e.g., "Summarize the upper player's tactic in this rally."). Critically, each question is paired with its corresponding Chain-of-Thought (CoT) rationale that simulates visual reasoning.

\paragraph{Knowledge-Based QA}
From an application perspective, this sub-task is designed to expand COACH's utility into a comprehensive badminton expert. It ensures the system can answer not only video-specific tactical questions but also general domain knowledge queries. This requirement, however, introduces a critical design challenge: how to handle these distinct query types efficiently. Using the full, multi-step visual collaboration pipeline (SOP) for a simple text-based question would be highly inefficient.
Therefore, it enriches the shared LLM backbone with specialized domain knowledge.

\subsubsection{Data Generation Pipelines}

\paragraph{Video-Based QA Generation}
Our pipeline for synthesizing the V-QA dataset begins by converting the structured stroke-level annotations from ShuttleSet (e.g., stroke type, player position) into dense, descriptive captions. These captions serve as the ground-truth visual context for the model. To ensure high quality and factual grounding, we employ an in-context learning approach. We manually authored a set of high-quality (Question, CoT, Answer) exemplars, which act as few-shot guides to constrain the reasoning style and output format of a 'Teacher' large language model (gpt-oss-120b~\cite{openai2025gptoss120bgptoss20bmodel}). In the final synthesis step, the Teacher LLM is prompted with both the ground-truth caption and the human-authored exemplars, and is instructed to generate both a complex tactical question and its corresponding detailed, step-by-step Chain-of-Thought (CoT) answer \cite{wei2023chainofthoughtpromptingelicitsreasoning}.

\paragraph{Knowledge-Based QA}
We crawled a comprehensive corpus of professional badminton knowledge from authoritative, publicly available web sources (e.g., \textit{BWF Statutes}, \textit{BadmintonBible}, \textit{wikipedia}, etc.). This corpus covers official rules, advanced techniques, footwork, and tactical principles. We then also used a teacher LLM to generate (Question, Answer) pairs based only on the provided textual context. All text sources were used for research purposes, adhering to fair use principles.

\subsection{Implementation Details}
\label{sec:implementation}
A key methodological challenge arises in establishing a fair comparison for our Generative Video Summarization application. This application is designed for long-form, often hour-plus, match videos. State-of-the-art end-to-end models, such as \textbf{Gemini 2.5 Pro} \cite{geminiteam2025geminifamilyhighlycapable}, are architecturally unsuited for processing such long-duration inputs while maintaining the fine-grained, stroke-level understanding required by our queries. Our COACH framework is explicitly designed to solve this by "chunking" the long-form video and dispatching these segments to the specialist \texttt{Grounder} agent for batch inference.

Therefore, to create a meaningful and direct comparison of the core capability required, our experiment is designed to evaluate the specialist \texttt{Grounder} agent against the generalist \textbf{Gemini 2.5 Pro} on their respective performance on these individual video chunks using low-level localization queries.

We compare our primary framework against a strong state-of-the-art baseline, \textbf{Gemini 2.5 Pro}, which represents the current frontier of end-to-end Video-Language Models (VLMs).

Our model adopts Flan-T5-XL \cite{chung2022scalinginstructionfinetunedlanguagemodels} as a single shared LLM backbone. Its vision components are based on TC-CLIP-B/16~\cite{kim2024leveragingtemporalcontextualizationvideo} as the video encoder and a Q-Former module for vision-language alignment. 

Before integration, these vision modules are aligned with the language space by captioning task. The entire framework is fully supervised trained on our \textbf{COACH-Dataset} for 2 epochs, using a learning rate of \texttt{2e-5} and a batch size of \texttt{16}.
\subsection{Evaluation Metrics}
\label{sec:metrics}
As our framework is designed as an application-focused generative system, we evaluate all models (COACH and baselines) on their natural language generation output. We employ a suite of specific metrics tailored to our different task categories to ensure a fair and rigorous comparison.

\paragraph{1. Factual Accuracy (EM)}
For tasks with a single definitive factual answer (e.g., Action Classification and Action Count), we use Exact Match (EM) Accuracy. We employ a parsing script to extract the core answer (e.g., the stroke name "smash" or the number "3") from the model's generated sentence and compare it to the ground truth.

\paragraph{2. Localization Fidelity (F1-Score)}
For all tasks requiring temporal localization , we evaluate the model's ability to precisely identify the correct event strokes. We extract the set of predicted stroke indices from the generated text and compare it to the ground-truth set, reporting the average Stroke-level F1-Score (along with Precision and Recall).

\paragraph{3. Hallucination Mitigation (NQA)}
To specifically measure the Grounder agent's robustness against hallucinations , we use Negative Query Accuracy (NQA). This metric is computed on the subset of localization queries where the event is absent, measuring the model's accuracy in correctly identifying these as non-existent (e.g., by reporting an empty set).

\paragraph{4. Generative Text Quality (ROUGE-L)}
For open-ended, text-only generative tasks (Summarisation and Knowledge QA), where semantic overlap is more important than exact match, we report the \textbf{ROUGE-L} score.
\subsection{Quantitative Results }
\label{sec:quantitative_results}

We present our main quantitative results in Table~\ref{tab:rally_qa_results} (Rally Video QA) and Table~\ref{tab:localization_results} (Scalable Temporal Grounding). Our analysis validates our core hypothesis: a specialist architecture, trained with fully supervised, structured CoT, is necessary for high-precision, domain-specific reasoning.

\subsubsection{Comparison with Generalist SOTA}
Our primary finding is the clear performance gap between our specialist \textbf{COACH} framework and the \textbf{Gemini 2.5 Pro} generalist baseline (Table~\ref{tab:rally_qa_results}).

This performance difference is most evident in high-precision tasks. In Temporal Localization, COACH achieves a 76.21\% F1-Score, a significant +62\% lead over Gemini (13.73\%). This result stems from our model's specialized training. Gemini's low precision (17.81\%) and recall (12.67\%) suggest its reasoning is statistically unreliable and not firmly "grounded" in the visual evidence, likely relying on generalized textual priors rather than the video's specific temporal structure.

Similarly, the +59\% and +42\% gains in Action Classification and Action Count, respectively, demonstrate the efficacy of our structured CoT pipeline. The Critic agent, enforcing a step-by-step verification process, enables COACH to successfully execute the discrete, factual reasoning required for these tasks—a capability that monolithic, end-to-end models are not optimized to perform.

It is indicated in the Scalable Temporal Grounding task (Table~\ref{tab:localization_results}). Here, the specialist Grounder agent (84.77\% F1) substantially outperforms the generalist Gemini (24.82\% F1). It is showing that a specialized architecture, trained via supervised CoT, is necessary for both high-level complex reasoning and low-level scalable grounding.

\subsubsection{Architectural Specialization vs. Model Scale}

The Knowledge QA task (Table~\ref{tab:rally_qa_results}) provides a critical point of analysis regarding model scale. On this text-only task, which primarily evaluates the raw knowledge of the LLM backbone, COACH (27.40 ROUGE-L) performs comparably to, but is slightly outperformed by, Gemini 2.5 Pro (29.76 ROUGE-L).

This result is consistent with our hypothesis. The performance gap on Knowledge QA  is expected, as the larger-scale Gemini model possesses superior pre-trained textual priors. This outcome effectively isolates the source of COACH's superior performance on all video-grounded tasks. It demonstrates that COACH's significant gains in temporal and factual reasoning are not a byproduct of superior model scale. Instead, this advantage is directly attributable to our specialized, agent-based architecture and the fully supervised, role-specific CoT patterns it was trained on.
\subsection{Ablation Studies}

\label{sec:ablations}
Having established the necessity of a specialist framework, we now conduct ablation studies to validate that COACH's high performance is a direct result of our key design decisions as described in Method.
\subsubsection{Impact of the Critic Agent}
To verify the importance of the Critic agent, we evaluate a variant, \textbf{COACH(w/o Critic)}, on the Analytical Rally QA task. It means the answers are directly output by Orchestrator agent without verification process.

As shown in Table~\ref{tab:rally_qa_results}, removing the Critic's verification step causes a significant 3.4\% drop in `Action Classification` accuracy (85.60\% $\to$ 82.20\%). This demonstrates that the Critic's evaluative, "backward-reasoning" mechanism is critical for mitigating factual errors and ensuring the reliability of the final answer.

\subsubsection{Validating Role-Specific Thinking Patterns}
To isolate the effect of the specialized thinking patterns learned via Structured CoT Tuning, we compare the performance of the Orchestrator agent against the specialist Grounder agent. Both agents were evaluated on the same Scalable Temporal Grounding task.

This comparison is particularly salient because both agents utilize the exact same shared backbone model (Flan-T5-XL). The sole difference between them is the distinct reasoning pattern (CoT) each was trained to follow. The Orchestrator employs the generalist, complex-reasoning pattern it learned, while the  Grounder  employs the stability-driven, "Observe $\to$ Report" pattern it was specialized for.

The results, presented in Table~\ref{tab:grounder_orchestrator_ablation}, are conclusive. The Grounder , by adhering to its specialized thinking pattern, significantly outperforms the Orchestrator (which uses its generalist pattern) on all metrics, most notably achieving a +8.82 F1-Score gain. Furthermore, its 91.80\% NQA score is 6.3 points higher, which demonstrates that its specialist pattern is significantly more effective at mitigating hallucinations by correctly identifying unanswerable queries. This experiment thus validates that our Structured CoT Tuning method can successfully create distinct, functionally specialized behaviors within a single model.

\begin{table}[htbp]
\centering
\caption{Performance of the specialist \textbf{COACH (\texttt{Grounder} Agent)} on the \textbf{Scalable Temporal Grounding} task (Task 2). We report metrics as percentages (\%).}
\label{tab:localization_results}
\begin{tabular}{|l|c|c|}
\hline
\textbf{Metric} & \textbf{Gemini 2.5 Pro} & \textbf{COACH} \\
\hline
hit@1 & 27.68 & \textbf{87.28} \\
EM & 15.53 & \textbf{72.31} \\
Precision & 27.49 & \textbf{86.95} \\
Recall & 25.16 & \textbf{84.65} \\
F1-Score & 24.82 & \textbf{84.77} \\
NQA & 21.23 & \textbf{91.80} \\
\hline
\end{tabular}
\end{table}
\begin{table}[htbp]
\centering
\caption{Ablation study validating the efficacy of \textbf{Role-Specific Thinking Patterns}. }
\label{tab:grounder_orchestrator_ablation}
\begin{tabular}{|l|c|c|}
\hline
\textbf{Metric} & \textbf{Orchestrator} & \textbf{Grounder} \\
\hline
Hit@1 & 82.73 & \textbf{87.28} \\
EM & 68.12 & \textbf{72.31} \\
Precision & 76.75 & \textbf{86.95} \\
Recall & 75.22 & \textbf{84.65} \\
F1-Score & 75.95 & \textbf{84.77} \\
NQA & 85.50 & \textbf{91.80} \\
\hline
\end{tabular}
\end{table}
\section{Conclusion}
\label{sec:conclusion}

In this paper, we proposed \textbf{COACH}, a composable, multi-agent framework built on a shared backbone \textbf{designed for system scalability}. We demonstrated that for specialized, high-stakes domains like sports analysis, a specialist architecture is necessary. Our experiments showed that COACH's specialist agents and verification mechanisms  achieve superior factual accuracy and grounding fidelity compared to state-of-the-art generalist models like Gemini 2.5 Pro. This architecture, which combines \textbf{Intent-Driven Orchestration} with \textbf{Structured CoT Tuning}, allows specialized agents to achieve high factual and temporal accuracy on tasks where generalist models fail.
\paragraph{Limitations}
Despite these promising results, our work has several limitations. First, our COACH-Dataset dataset, while comprehensive for badminton, is currently limited to a single sport. Second, our policy-guided orchestration  relies on a fixed set of pre-defined SOPs; this ensures stability but limits the system's flexibility in handling completely novel, unseen task structures that do not fit the QA or Localization paradigms.Third, the evaluation of the Video Summarization pipeline was  focused on the quantitative of temporal localization (e.g., F1, EM) to validate our Grounder agent. We acknowledge that a "good" summary involves more than just accurate localization, including qualitative aspects such as narrative coherence, interestingness, and contextual relevance.

\paragraph{Future Work}
These limitations highlight clear avenues for future research. Our immediate next step is to address the holistic evaluation of summarization. We plan to conduct a \textbf{comprehensive human evaluation (user study)} to assess the \textit{end-to-end qualitative performance} of the generated video summaries. Concurrently, we will leverage the framework's composability to add new SOPs for high-level generative tasks, such as generating \textbf{journalistic press releases} or \textbf{in-depth tactical analysis reports} for coaches.
\section{Acknowledgments}
This work was supported by the Ministry of Science and
Technology of Taiwan under Grants NSTC-113-2425-H-A49-001.
\bibliography{aaai2026}


\end{document}